# Inexpensive and Portable System for Dexterous High-Density Myoelectric Control of Multiarticulate Prostheses

Jacob A. George, *Student Member, IEEE,* Sridharan Radhakrishnan, Mark R. Brinton, and Gregory A. Clark

*Abstract*— Multiarticulate bionic arms are now capable of mimicking the endogenous movements of the human hand. 3D-printing has reduced the cost of prosthetic hands themselves, but there is currently no low-cost alternative to dexterous electromyographic (EMG) control systems. To address this need, we developed an inexpensive (~$675) and portable EMG control system by integrating low-cost microcontrollers with an EMG acquisition device. We validated signal acquisition by comparing the signal-to-noise ratio (SNR) of our system with that of a high-end research-grade system. We also demonstrate the ability to use the low-cost control system for proportional and independent control of various prosthetic hands in real-time. We found that the SNR of the low-cost control system was statistically no worse than 44% of the SNR of a research-grade control system. The RMSEs of predicted hand movements (from a modified Kalman filter) were typically a few percent better than, and not more than 6% worse than, RMSEs of a research-grade system for up to six degrees of freedom when only relatively few (six) EMG electrodes were used. However, RMSEs were generally higher than RMSEs of research-grade systems that utilize considerably more (32) EMG electrodes, guiding future work towards increasing electrode count. Successful instantiation of this low-cost control system constitutes an important step towards the commercialization and wide-spread availability of dexterous bionic hands.

## I. INTRODUCTION

More than 1.6 million individuals in the United States suffer from limb-loss [1], which leads to a chronic struggle with pain, depression, and functional disability [1], [2]. On top of this, the high cost of upper-limb prostheses places financial strain on the limb-loss community [3]. Up to 50% of upper-limb amputees abandon or limit prosthesis use [4] due to ineffective control and high cost (e.g., of repairs) [5].

3D-printing has substantially reduced the price of multiarticulate prosthetic hands [6], but the ability to simultaneously and independently control the many degrees of freedom (DOFs) on these hands remains costly and out of reach for most amputees. Dexterous control strategies often employ high-density electromyography (EMG) [7] and utilize more computationally expensive machine-learning algorithms such as Kalman filters [8], [9] and neural networks [10]. Traditionally, these algorithms are instantiated on desktop computers or portable research-grade systems that can cost up to $64,000.

Here, we describe the development and validation of an inexpensive control system that can bring simultaneous and proportional control to 3D-printed prosthetics to eliminate the financial barriers associated with dexterous prostheses. We first describe the system design and material costs, and then demonstrate comparable signal acquisition and dexterous control to that of a high-end research-grade system. Altogether, this work highlights that dexterous control algorithms can be readily instantiated on low-cost control systems, thereby increasing the availability of dexterous prostheses to amputees and researchers alike.

## II. DEVICE OVERVIEW

### A. Design Criteria

The overall design objective was to develop an inexpensive and portable control system capable of providing independent and proportional control over three or more DOFs in real-time from high-density EMG. These criteria were established *a priori* based on currently available multiarticulate prostheses, dexterous control algorithms and research-grade control systems.

### B. Low-Cost Design

The control system consists of five major components (Fig. 1). These are: 1) electrodes for recording high-density surface EMG; 2) circuitry for amplifying and filtering EMG (Muscle SpikerShield Pro; Backyard Brains, Ann Arbor, MI, USA); 3) a microcontroller for sampling EMG data (Mega 2560; Arduino, Somerville, MA, USA); 4) a minicomputer for implementing dexterous control algorithms (Raspberry Pi 3b+; Raspberry Pi Foundation, Cambridge, UK); and 5) an external battery (PowerCore 2000 Redux; Anker, Shenzhen, China). These components minimize the total cost (~$675) of the system while still meeting the design criteria (Table 1).

Table 1. Itemized Expenses for Low-Cost Control System

| ITEM | PURPOSE | COST |
|---|---|---|
| Muscle SpikerShield Pro | Signal filtering & amplifying | $399.99 |
| Arduino Mega 2560 | Signal acquisition | $35.50 |
| Raspberry Pi 3b+ | Control algorithms | $45.99 |
| 32-GB Sandisk SD card | Data storage | $19.99 |
| PowerCore 20000 Redux portable power bank | Portable battery | $49.99 |
| Other materials (cables, buttons, electrodes, etc.) | Interfacing | ~$60.00 |
| 3D-printing | Portable case | ~$65.00 |
| **TOTAL:** | | **~$675.00** |

### C. High-Density EMG Recordings

Conventional myoelectric prostheses use two bipolar input channels that broadly target flexors and extensors to provide independent, but binary, control of two DOFs [11], [12].

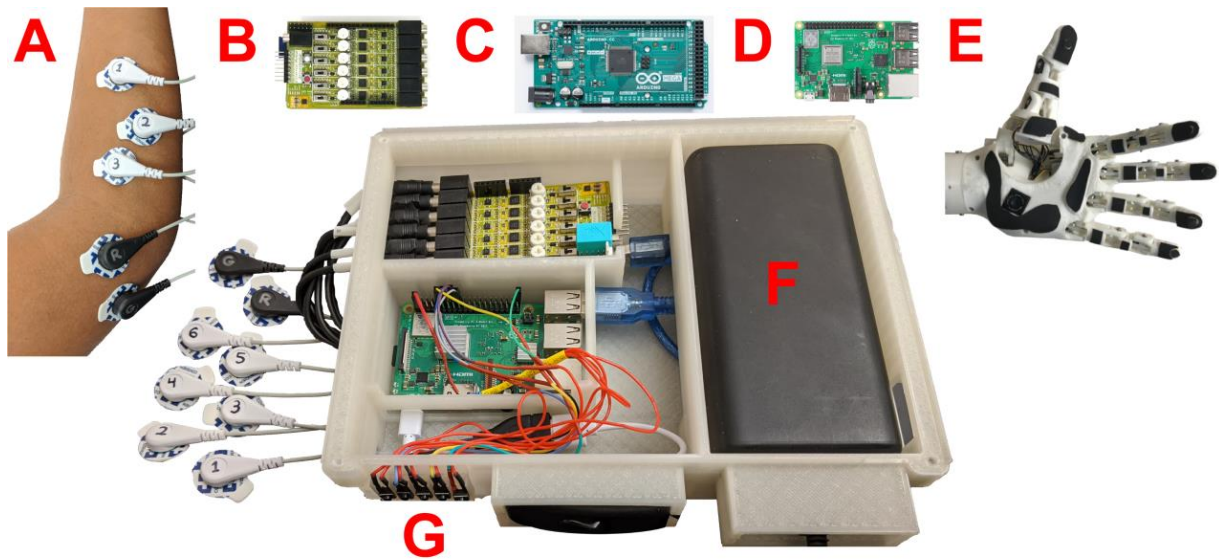

Figure 1. Overview of the low-cost control system. **A)** Six channels of electromyographic (EMG) data are acquired from the extrinsic hand muscles in the forearm using snap electrodes. **B)** Signals are filtered and amplified using a Muscle SpikerShield Pro from Backyard Brains. **C)** Data are sampled at 1 kHz using an Arduino Mega microcontroller. **D)** A Raspberry Pi minicomputer is used to calculate EMG features. **E)** A modified Kalman filter embedded onto the minicomputer predicts hand kinematics in order to control low-cost 3D-printed prostheses in real-time (e.g., the HANDI Hand). **F)** The system is completely portable and an external battery provides 8+ hours of use. **G)** Buttons embedded on the outside of the system allow users to train and run multiple dexterous myoelectric control algorithms.

Proportional control of three or more DOFs often uses more than 2 channels [9], [10], [13]. We incorporated six bipolar input channels in the system using the Muscle SpikerShield Pro. To further expand the electrode density, we used a common reference and ground for all bipolar inputs and calculated all possible differential pairs (six choose two) at 1 kHz [9] using the microcontroller. The system is comprised of eight surface electrodes (six single-ended recording electrodes, one reference electrode and one ground electrode) that yield a total of 21 EMG channels (six single-ended recordings plus 15 unique differential pairs) sampled at 1 kHz. EMG signals were amplified and band-pass filtered with cutoff frequencies of 55 Hz and 2500 Hz, and low-pass filtered with cutoff frequency of 3000 Hz. The filtered EMG data was then transmitted to the Raspberry Pi via serial communication for further feature extraction.

### D. EMG Feature Extraction

Various forms of feature extraction can be implemented and performed in real-time on the minicomputer. We demonstrate the ability to calculate the mean absolute value in real-time (a common EMG feature [9]). The mean absolute value was smoothed using an overlapping 300-ms window. The resulting EMG feature set consisted of the 300-ms smoothed mean absolute value on 21 channels, calculated at ~25 Hz.

### E. Data Collection and Storage

Decoding motor intent from EMG activity depends on the ability to correlate EMG features (e.g., the mean absolute value) to intended hand kinematics. Participants were instructed to mimic preprogrammed movements of the prosthetic hand. Synchronized EMG features, raw EMG data, and kinematics were written to CSV files at ~25 Hz. Data were stored locally using an expandable SD card. A 32-GB SD card was capable of storing ~108,000 minutes of training data, which is 10,800 times greater than traditionally used [9].

### F. Proportional and Independent Control

We implemented a modified Kalman filter [9] to demonstrate real-time proportional and independent control of more than three DOFs. The baseline mean absolute values were subtracted from the EMG features prior to training and testing the modified Kalman filter. We bound the output of the Kalman filter between -1 and 1 to match the control limits of various 3D-printed prostheses, such that -1 corresponded to maximum extension/adduction/supination, +1 corresponded to maximum flexion/abduction/pronation, and the hand was at rest at zero.

### G. Portability & Packaging

The entire low-cost control system was assembled and packaged into a custom 3D-printed case (19.37 x 22.9 x 5.4 cm; Fig. 1). A 4.8-Amp power bank was used to power the system, providing a maximum run-time capacity of ~8 hours. Two pairs of buttons were embedded into the 3D-printed case to initiate the training sequence and run-time control for two different control algorithms. A custom SAMTEC cable provided quick and modular attachments to various surface electrode configurations.

## III. METHODS

### A. Validation of Signal Acquisition

To validate signal acquisition, we measured the signal-to-noise ratio (SNR) of the low-cost control system relative to a higher-end research-grade control system (Grapevine Neural Interface Processor; Ripple Neuro LLC, Salt Lake City, UT USA). This higher-end research-grade control system utilized a band-pass filter with cutoff frequencies of 15 Hz and 375

Hz, and notch filters at 60, 120 and 180 Hz. For three able-bodied participants, six single-ended surface electrodes were placed on the forearm, with three electrodes targeting the extrinsic hand flexors and three electrodes targeting the extrinsic hand extensors. A common reference and ground were placed proximal to the elbow. Electrodes were wired directly to both the low-cost control system and the research-grade control system.

Participants were instructed to repeatedly perform three different sets of movements for ten seconds each: 1) individual digit movements (isolated flexion and extension of D1-D5), 2) grasping (simultaneous flexion and extension of D1-D5), and 3) wrist movement (e.g., rotation, flexion/extension, and deviation). This was followed by a 10-second rest period with no muscle activity. Each of the three participants repeated this 40-second data acquisition three times. Data was recorded by and synchronized between the low-cost control system and research-grade control system.

The signal-to-noise ratio, defined as the mean absolute value during the movement divided by the mean absolute value during rest, was then calculated for each of the three movement types. For each movement type, a two one-sided t-test for equivalence (TOST, [14]) was used to determine the minimum equivalence interval for which the electrode recordings from the low-cost control system and research-grade control system were statistically equivalent (with $\alpha = 0.05$ and $N = 18$ electrodes).

### B. Offline Analysis of Independent and Proportional Control

Using the same surface electrode configuration described above, six participants (different from the three participants used for the SNR comparison) were instructed to mimic preprogrammed movements of a 3D-printed prosthetic hand (HANDI Hand; BLINC Labs [6]). These movements included abduction/adduction of D1 and individual flexion/extension of D1–D5, for a total of 6 movements. Synchronized EMG features and kinematics were recorded in real-time and saved to a CSV file stored locally on the low-cost control system. This data was used to train a modified Kalman filter which then ran in real-time.

To assess proportional control, we quantified the Root Mean Squared Error (RMSE) of intended movements for the modified Kalman filter when trained on a random 50% of the data and tested on the remaining 50% of the data. Likewise, to access independent control of the DOFs, we quantified RMSE of unintended movements. These metrics have been used before for this algorithm [9], [10].

Performance was evaluated for a single controllable DOF up to six controllable DOFs. All possible combinations of DOFs were explored for all participants and then averaged. For example, the value reported for a single DOF was the average of six possible DOF combinations, the value reported for two DOFs was the average of 15 possible DOF combinations (six choose two), etc.

For each number of controllable DOFs, a two one-sided test for equivalence (TOST) was used to determine the minimum equivalence interval for which the RMSEs from the low-cost control system and research-grade control system were statistically equivalent (with $\alpha = 0.05$ and $N = 6$ participants).

## IV. RESULTS

### A. Portable Low-Cost Control System Provides Dexterous Myoelectric Control in Real-Time

The low-cost control system meets the established design criteria and reduces the total cost of dexterous prosthetic control by two orders of magnitude (Table 2). With a total cost of ~$675, this low-cost control system can store 1800+ hours of EMG activity from 21 unique channels in a completely portable formfactor with 8+ hours of battery life. Furthermore, the system provides enough processing power for real-time signal acquisition, data logging, kinematic prediction, and prosthetic control. The system is adaptable for various control algorithms and prosthetic devices, and is scalable for increased battery life, data storage, or processing power.

Table 2. Design Criteria and Specifications

| GOAL | ACTUAL |
|---|---|
| Inexpensive | $675 |
| Real-time control | Fixed 40-ms updated speed |
| Store large datasets of synchronized EMG and kinematics | Expandable storage capabilities; 1800+ hours of synchronized EMG/kinematic data |
| High-density EMG recordings | Records 21 unique EMG channels |
| Dexterous control of multi-articulate prostheses | Modified Kalman filter provides proportional and independent control of 6 DOFs |
| Portable take-home system | Worn on the hip or carried in a backpack; 8 hours of battery life |
| Adaptable to various terminal devices | Serial communication established. Bluetooth and CAN communication possible. |

### B. SNR Was Not Significantly Different between Low-Cost and High-Cost Systems

To validate signal-acquisition capabilities, we compared the SNR of the low-cost control system to that of a high-end research-grade control system, respectively, for three different movement patterns. We found that the SNRs (mean ± standard deviation) of the two systems were comparable during individual digit movements (2.08 ± 0.83 vs 2.38 ± 1.67), grasping (3.69 ± 0.99 vs 4.51 ± 2.78), and wrist movements (2.90 ± 0.72 vs 3.34 ± 1.90) (Fig. 2A). The SNR of the low-cost system was statistically equivalent to that of the research-grade system within +0.45 or -1.05 SNR for individual digit movements, within +0.36, -1.99 SNR for grasping, and within +0.38, -1.25 SNR for wrist movements ($p$'s < 0.05, TOST). In other words, the SNR of the low-cost control system is statistically no more than 44% worse than the high-end research-grade system for individual digit movements or for grasping, and no more than 37% worse for wrist movements.

### C. Modified Kalman filter Can Be Run in Real-Time to Provide Independent and Proportional Control

We implemented a modified Kalman filter [9] onto the low-cost control system in order to achieve independent and proportional control of six degrees of freedom in real-time.

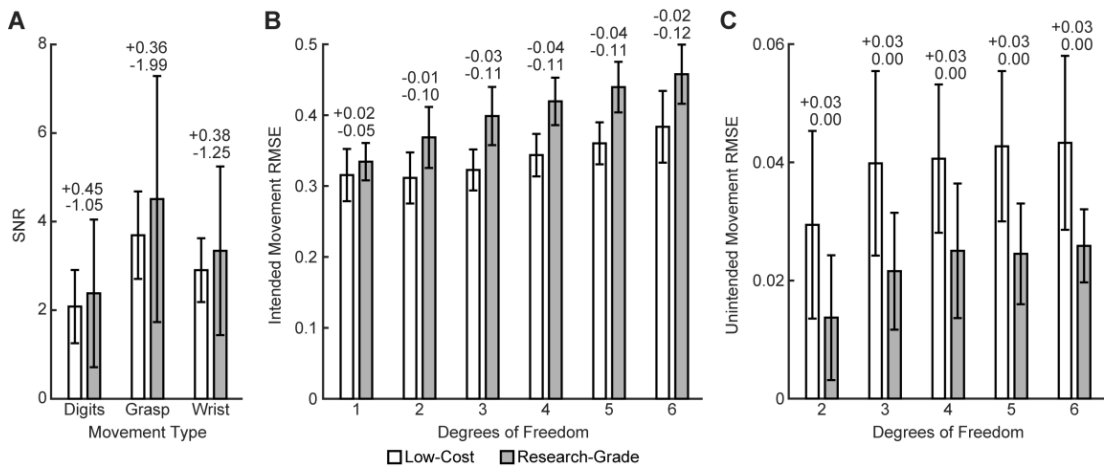

Figure 2. Low-cost control system reduces the total cost of dexterous myoelectric control by two orders of magnitude with relatively little decrease in performance, at least for systems with relatively few EMG electrodes. Bars show mean ± standard deviation. Values above the bars represent the upper (top) and lower (bottom) statistical equivalence bounds, relative to the research-grade system ($p$'s < 0.05, two one-sided t-tests for equivalence). **A)** The signal-to-noise ratio (SNR) of the low-cost control system was statistically no worse than that of the research-grade system within an equivalence window of 44% for digits (100 x -1.05/2.38) or for grasp (100 x -1.99/4.51), or 37% for wrist (100 x -1.25/3.34). **B)** A modified Kalman filter was implemented on the low-cost control system to control up to six degrees of freedom (DOFs) of a prosthetic hand independently and proportionally in real-time. Low RMSEs indicate better performance. For one controllable DOF, the RMSE of intended movements of the low-cost control system was statistically no more than 6% worse than the RMSE of the research-grade system (100 x 0.02/0.33). For six controllable DOFs, the RMSE of the low-cost system was statistically at least 4% better than the RMSE of the research-grade system (100 x -0.02/0.46). **C)** The RMSEs of unintended movements of the low-cost control system were statistically no greater than 0.03 larger than the corresponding RSMEs of the research-grade system for one to six controllable DOFs.

The RMSE of intended movements for the low-cost control system was statistically no more than 6% worse than the RMSE of the research-grade system for one controllable DOF, and was statistically at least 4% better than the research-grade system RMSE for six controllable DOFs ($p$'s < 0.05, TOST; Fig. 2B). The RMSEs of unintended movements of the low-cost control system were statistically no greater than 0.03 larger than the corresponding RSMEs of the research-grade system for one to six controllable DOFs ($p$'s < 0.05, TOST; Fig. 2C). Differences between RMSEs of intended and unintended movements may be due to slight variations in training data or differences in filtering capabilities.

### D. Portable Real-Time Control for Activities of Daily Living

Participants were able to use the low-cost control system to intuitively control a six-DOF prosthetic hand in real-time (Fig. 3). Although not formally tested, participants were able to use the prosthesis alone or in conjunction with their intact hand to manipulate fragile objects and shake hands with themselves.

## V. DISCUSSION

This work highlights the development of an inexpensive and portable control system capable of providing independent and proportional control of six DOFs in real-time from high-density EMG. Coupled with recent advancements in low-cost 3D-printed multiarticulate prostheses, the successful instantiation of this low-cost control system constitutes an important step towards the commercialization and wide-spread availability of low-cost dexterous bionic hands.

This work builds from prior research on low-cost prostheses with integrated myoelectric control [15]–[17] by introducing high-density EMG and a more dexterous control algorithm. Prior low-cost implementations utilized only a single channel of EMG [15], controlled only a single DOF [16], or provided only binary control [17]. In the present study, we extend these previous demonstrations to proportional and independent control of multiarticulate prostheses.

The results with a research-grade control system presented here, in comparison to what has been published previously, suggest that more than six EMG electrodes are necessary for optimal performance—there is roughly a 40% reduction in intended movement RMSE when 32 electrodes were used instead of only six [9]. Thus, although no direct empirical comparisons were made in the present study, the performance of the research-grade system using considerably more EMG electrodes appears to be superior to the performance of the low-cost system here using only six EMG electrodes. Additionally, more than 50% of the total cost of the low-cost control system was dedicated to circuitry for amplifying and filtering six EMG electrodes. Future work should leverage low-cost analog-to-digital converters (ADS1298; Texas Instruments, Dallas, USA) to increase electrode count and lower the overall cost.

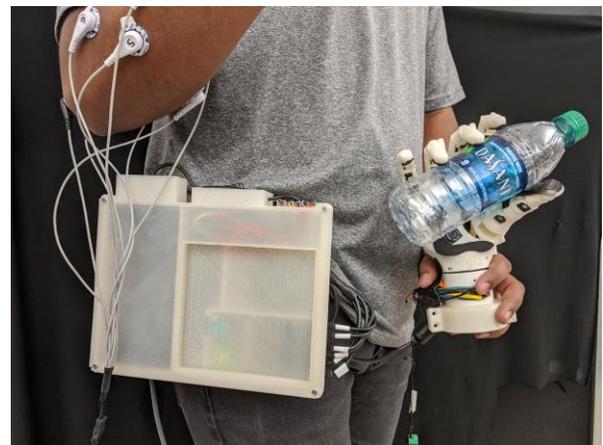

Figure 3. Low-cost control system for use in activities of daily living. Participants were able to use the portable low-cost control system to perform several one- and two-handed tasks. One participant is shown grasping a water bottle with an inexpensive prosthesis controlled by a modified Kalman filter and the low-cost control system.

This work also highlights the computational efficiency of the modified Kalman filter. With a run-time of ~40 ms, participants can control prostheses in real time. However, recent trends towards deep-learning for myoelectric control suggest that more computationally demanding approaches may yield more robust control [18]. To this end, future work should integrate low-cost control systems with inexpensive TPU accelerators (Coral USB Accelerator; Google LLC, Mountain View, USA). Freeing up additional computational time will also enable pathways for integrating low-cost sensory feedback for closed-loop control [19], [20].

## VI. Conclusion

As 3D-printed bionic arms become more affordable and dexterous, there is a parallel need to make control systems more affordable and dexterous as well. This work demonstrates that low-electrode count EMG and advanced algorithms for proportional and independent control of multiple DOFs can be readily implemented on portable and low-cost components. At ~$675, the low-cost control system presented here provides an immediate avenue to increase the availability of dexterous prostheses to amputees and researchers alike.

## Author Contributions

J.A.G. designed experiments, designed algorithms, collected data, performed data analysis, wrote the manuscript and oversaw device design. S.R. designed the devices, implemented software, performed experiments, assisted with data analysis, and helped draft the manuscript. M.R.B. assisted with designing and troubleshooting the devices. G.A.C. oversaw all aspects of the research. All authors contributed to the revision of the manuscript.

## Acknowledgment

This work was funded by: DARPA, BTO, Hand Proprioception and Touch Interfaces program, Space and Naval Warfare Systems Center, Pacific, Contract No. N66001-15-C-4017; NSF Award No. ECCS-1533649; NSF Award No. CHS-1901492 and NSF Award No. GRFP-1747505.

## References

[1] K. Ziegler-Graham, E. J. MacKenzie, P. L. Ephraim, T. G. Travison, and R. Brookmeyer, "Estimating the prevalence of limb loss in the united states: 2005 to 2050," *Arch. Phys. Med. Rehabil.*, vol. 89, no. 3, pp. 422–429, Mar. 2008, doi: 10.1016/j.apmr.2007.11.005.

[2] C. G. Bhuvaneswar, L. A. Epstein, and T. A. Stern, "Reactions to amputation: recognition and treatment," *Prim. Care Companion J. Clin. Psychiatry*, vol. 9, no. 4, pp. 303–308, 2007.

[3] D. K. Blough, S. Hubbard, L. V. McFarland, D. G. . Smith, J. M. Gambel, and G. E. Reiber, "Prosthetic cost projections for servicemembers with major limb loss from Vietnam and OIF/OEF," *J. Rehabil. Res. Dev.*, vol. 47, no. 4, p. 387, 2010, doi: 10.1682/JRRD.2009.04.0037.

[4] E. A. Biddiss and T. T. Chau, "Upper limb prosthesis use and abandonment: A survey of the last 25 years," *Prosthet. Orthot. Int.*, vol. 31, no. 3, pp. 236–257, Sep. 2007, doi: 10.1080/03093640600994581.

[5] E. Biddiss and T. Chau, "Upper-limb prosthetics: critical factors in device abandonment," *Am. J. Phys. Med. Rehabil.*, vol. 86, no. 12, pp. 977–987, Dec. 2007, doi: 10.1097/PHM.0b013e3181587f6c.

[6] D. J. A. Brenneis, M. R. Dawson, and P. M. Pilarski, "Development of the HANDi Hand: An Inexpensive, Multi-Articulating, Sensorized Hand for Machine Learning Research in Myoelectric Control," in *Myoelectric Controls and Upper Limb Prosthetics Symposium*, Fredericton, New Brunswick, Canada, 2017, p. 4.

[7] A. Matran-Fernandez, I. J. Rodríguez Martínez, R. Poli, C. Cipriani, and L. Citi, "SEEDS, simultaneous recordings of high-density EMG and finger joint angles during multiple hand movements," *Sci. Data*, vol. 6, no. 1, pp. 1–10, Sep. 2019, doi: 10.1038/s41597-019-0200-9.

[8] J. Nieveen *et al.*, "Polynomial Kalman filter for myoelectric prosthetics using efficient kernel ridge regression," in *2017 8th International IEEE/EMBS Conference on Neural Engineering (NER)*, 2017, pp. 432–435, doi: 10.1109/NER.2017.8008382.

[9] J. A. George, T. S. Davis, M. R. Brinton, and G. A. Clark, "Intuitive neuromyoelectric control of a dexterous bionic arm using a modified Kalman filter," *J. Neurosci. Methods*, vol. 330, p. 108462, Nov. 2019, doi: 10.1016/j.jneumeth.2019.108462.

[10] J. A. George, M. R. Brinton, C. C. Duncan, D. T. Hutchinson, and G. A. Clark, "Improved Training Paradigms and Motor-decode Algorithms: Results from Intact Individuals and a Recent Transradial Amputee with Prior Complex Regional Pain Syndrome," in *2018 40th Annual International Conference of the IEEE Engineering in Medicine and Biology Society (EMBC)*, 2018, pp. 3782–3787, doi: 10.1109/EMBC.2018.8513342.

[11] M. R. Dawson, F. Fahimi, and J. P. Carey, "The Development of a Myoelectric Training Tool for Above-Elbow Amputees," *Open Biomed. Eng. J.*, vol. 6, pp. 5–15, Feb. 2012, doi: 10.2174/1874230001206010005.

[12] P. Geethanjali, "Myoelectric control of prosthetic hands: state-of-the-art review," *Med. Devices Auckl. NZ*, vol. 9, pp. 247–255, Jul. 2016, doi: 10.2147/MDER.S91102.

[13] N. Jiang and D. Farina, "Myoelectric control of upper limb prosthesis: current status, challenges and recent advances," *Front. Neuroengineering*, doi: 10.3389/conf.fneng.2014.11.00004.

[14] J. L. Rogers, K. I. Howard, and J. T. Vessey, "Using significance tests to evaluate equivalence between two experimental groups," *Psychol. Bull.*, vol. 113, no. 3, pp. 553–565, May 1993, doi: 10.1037/0033-2909.113.3.553.

[15] I. Ku, G. K. Lee, C. Y. Park, J. Lee, and E. Jeong, "Clinical outcomes of a low-cost single-channel myoelectric-interface three-dimensional hand prosthesis," *Arch. Plast. Surg.*, vol. 46, no. 4, pp. 303–310, Jul. 2019, doi: 10.5999/aps.2018.01375.

[16] N. Sreenivasan, D. F. Ulloa Gutierrez, P. Bifulco, M. Cesarelli, U. Gunawardana, and G. D. Gargiulo, "Towards Ultra Low-Cost Myoactivated Prostheses," *BioMed Res. Int.*, vol. 2018, Oct. 2018, doi: 10.1155/2018/9634184.

[17] M. M. Atique and S. Rabbani, "A Cost-Effective Myoelectric Prosthetic Hand," *JPO J. Prosthet. Orthot.*, vol. 30, no. 4, p. 231, Oct. 2018, doi: 10.1097/JPO.0000000000000211.

[18] H. Dantas, D. J. Warren, S. Wendelken, T. Davis, G. A. Clark, and V. J. Mathews, "Deep Learning Movement Intent Decoders Trained with Dataset Aggregation for Prosthetic Limb Control," *IEEE Trans. Biomed. Eng.*, pp. 1–1, 2019, doi: 10.1109/TBME.2019.2901882.

[19] C. Pylatiuk, A. Kargov, and S. Schulz, "Design and Evaluation of a Low-Cost Force Feedback System for Myoelectric Prosthetic Hands:," *JPO J. Prosthet. Orthot.*, vol. 18, no. 2, pp. 57–61, Apr. 2006, doi: 10.1097/00008526-200604000-00007.

[20] K. R. Schoepp, M. R. Dawson, J. S. Schofield, J. P. Carey, and J. S. Hebert, "Design and Integration of an Inexpensive Wearable Mechanotactile Feedback System for Myoelectric Prostheses," *IEEE J. Transl. Eng. Health Med.*, vol. 6, pp. 1–11, 2018, doi: 10.1109/JTEHM.2018.2866105.